\documentclass[11pt,a4,onecolumn]{article}  %
\usepackage[margin=1in,footskip=0.25in]{geometry}
\usepackage{amsmath,amsfonts,amssymb}
\usepackage{graphicx, subfigure}
\usepackage{setspace}
\usepackage[subfigure]{tocloft}
\usepackage{enumitem}

\usepackage{multirow} 
\usepackage{setspace} 
\usepackage[ruled,vlined,linesnumbered]{algorithm2e}
\SetKwComment{Comment}{$\triangleright$\ }{}
\usepackage{threeparttable} 
\usepackage{xcolor} 
\usepackage{adjustbox} 

\usepackage{mathtools} 
\usepackage{authblk} 
\usepackage{multicol}
\usepackage{float}
\usepackage{tabularx} 

\begin{document}
\title{Adaptive strategy for superpixel-based region-growing image segmentation} 
\author[a,b*]{Mahaman Sani Chaibou}
\author[b,c]{Pierre-Henri Conze}
\author[a]{Karim Kalti}
\author[b]{Basel Solaiman}
\author[a]{Mohamed Ali Mahjoub}
\affil[a]{LATIS, ENISo, Sousse University, 264 Erriadh, Sousse 4023, Tunisia}
\affil[b]{Institut Mines-T{\'e}l{\'e}com Atlantique, Technop{\^o}le Brest-Iroise, CS 83818, 29238 Brest Cedex 03, France}
\affil[c]{LaTIM UMR 1101, INSERM, 2 avenue Foch, 29609 Brest Cedex, France}

\renewcommand{\cftdotsep}{\cftnodots}
\cftpagenumbersoff{figure}
\cftpagenumbersoff{table} 

\maketitle

\begin{abstract}
This work presents a region-growing image segmentation approach based on superpixel decomposition.
From an initial contour-constrained over-segmentation of the input image, the image segmentation is achieved by iteratively merging similar superpixels into regions. This approach raises two key issues: (1) how to compute the similarity between superpixels in order to perform accurate merging and (2) in which order those superpixels must be merged together. 
In this perspective, we firstly introduce a robust adaptive multi-scale superpixel similarity in which region comparisons are made both at content and common border level. Secondly, we propose a global merging strategy to efficiently guide the region merging process. Such strategy uses an adpative merging criterion to ensure that best region aggregations are given highest priorities. This allows to reach a final segmentation into consistent regions with strong boundary adherence.
We perform experiments on the BSDS500 image dataset to highlight to which extent our method compares favorably against other well-known image segmentation algorithms. The obtained results demonstrate the promising potential of the proposed approach.

\end{abstract}



\begin{spacing}{1.5}   

\section{Introduction}\label{sect:intro}  
Image segmentation is a fundamental task in many pattern recognition and computer vision applications such as object detection, content-based image retrieval and medical image analysis. Segmentation is the process that consists of partitioning an image into homogeneous regions of pixels with similar characteristics and spatially accurate boundaries\cite{haralick1985image}.
Despite the simplicity of its definition, image segmentation is a hard problem that does not have a universal solution. Beside, one should not expect the segmentation of an image to be unique \cite{yang2008unsupervised} because of at least two reasons, according to Zhuowen\cite{tu2002image}: (1) it is fundamentally complex to model the vast amount of visual patterns of images, (2) perception is intrinsically ambiguous. Indeed, quite often one can provide different logical interpretations for the same image.
Image segmentation is a field that has been extensively studied for decades and the existing methods can be classified into clustering-based, boundary-based and region-based\cite{adams1994seeded,shih2005automatic}. 
Clustering-based methods segment images by classifying pixels based on their extracted properties\cite{conze2016,zhang2006adaptive,arora2008multilevel,chen2016multi}.
Boundary-based methods use the assumption that pixels properties change abruptly between different regions.\cite{senthilkumaran2009edge,arbelaez2011contour,dollar2013structured,zhao2017optimal} 
Region-based methods assume that a region is composed of adjacent pixels with similar properties\cite{baazaoui2017semi,castillo2016k,zhu2016medical,hu2017brain}.
There are some hybrid approaches that combine two or more of the aforementioned methods\cite{hettiarachchi2017voronoi}.

Region-growing is a popular region-based segmentation technique that operates by merging regions with similar pixels on their borders in an iterative fashion. At each iteration, all pixels that border the growing region are examined and the most similar are appended to that region. Initial regions may be pixels or regions produced by dedicated over-segmentation techniques, in which case they are called superpixel\cite{ren2003learning}. A superpixel is commonly defined as a perceptually meaningful atomic region obtained by aggregating neighboring pixels based on spatial and appearance similarity criteria. 
In recent years, superpixel-based image segmentation techniques have gained a big interest among the image processing community mainly for their computational efficiency. Superpixels also allow more efficient semantic feature extraction contrary to image patches\cite{machairas2016new}. 
However, those techniques present two main issues: the similarity measure between superpixels and the superpixels dependencies. 
The similarity measure refers to the quantification of how similar two superpixels are. This is usually computed by a normalized distance between the superpixels.
Mehnert and Jackway\cite{mehnert1997improved} stated that a region-growing algorithm is inherently dependent on the processing order of the image pixels. Either, whenever several pixels have the same distance from their neighboring pixels or whenever one pixel has the same distance from several regions.
\\
As a solution for the aforementioned issues, we propose in this work a robust similarity measure between superpixels as well as an adaptive superpixel merging strategy. The similarity measure integrates content and border information jointly to provide robust similarity between superpixels. The merging strategy uses an auto updated criterion to iteratively aggregate superpixels based on the proposed similarity measure.
This paper is organized as follows. Section~\ref{sect:relatedworks} presents a literature analysis related to the investigated segmentation technique. The proposed approach is fully detailed in Section~\ref{sect:method} including a complete overview (Sect.\ref{sect:overview}) as well as a focus on superpixel and region features (Sect.\ref{sect:SpClustersfeatures}), region similarity measures (Sect.\ref{sect:ClustersSimilarity}) and region aggregation strategy (Sect.\ref{sect:ConsistencyBC}). An extension of the well-known over-segmentation algorithm SLIC is proposed in Section~\ref{sect:ConstrainedSLIC}.
We provide in Section \ref{sect:results} an experimental assessment of our method in comparison with state-of-the-art segmentation algorithms. Finally, Section \ref{sect:conclusion} concludes our study.

\section{Related work}\label{sect:relatedworks}
\subsection{Simple Linear Iterative Clustering}
A superpixel decomposition of an image consists of a partition of the image into small perceptually meaningful regions. They provide a handy representation of the image that heavily reduces the number of visual primitives. According to Stutz et al.\cite{StutzHermansLeibe2016}, any superpixels over-segmentation algorithm should fall in one of the seven following categories: watershed-based, density-based, graph-based, contour-based, path-based, clustering-based and energy optimization. 
Simple Linear Iterative Clustering (SLIC)\cite{achanta2010slic} is a clustering-based algorithm and one of the most commonly used to generate superpixels. It offers a simple implementation and provides compact and nearly uniform superpixels. The cluster centers are initialized on a uniform grid and pixels in a $2S\times2S$ window around clusters are iteratively aggregated according to the metric $d$ (eq.\ref{eqn:slicdist}) defined in a five-dimensional space composed by three features and two spatial components.
\begin{equation}\label{eqn:slicdist}
d=\sqrt{\left(\dfrac{d_c}{m}\right)^2+\left(\dfrac{d_S}{S}\right)^2}
\end{equation}
where $d_c=\sqrt{(l_j-l_i)^2+(a_j-a_i)^2+(b_j-b_i)^2}$ is the distance between the color feature vectors and $d_S=\sqrt{(x_j-x_i)^2+(y_j-y_i)^2}$ is the distance between the spatial coordinate vectors of the current cluster center $C_k$ and the considered neihbor pixel $i$. $m$ is a weighting term used to control regularity of generated superpixels and the initial cluster centers interval $S=\sqrt{N/K}$ is the size of the neighborhood window around cluster centers; with $N$ being the total number of pixel in the image and $K$ the desired number of approximately equally-sized superpixels.
A more comprehensive survey on superpixels image over-segmentation algorithms can be found in the work of Stutz and al.~\cite{StutzHermansLeibe2016}.

\subsection{Superpixel-based image segmentation}
Superpixels over-segmentations are usually used as an initilization for image segmentation for two main reasons\cite{yin2017unsupervised}. Firstly, they generally adhere to boundaries and produce meaningful small regions. Secondly, they drastically reduce computation time by reducing the number of processed elements. 
Hsu and Ding \cite{hsu2013efficient} proposed a segmentation approach for natural images that uses Simple Linear Iterative Clustering (SLIC)\cite{achanta2010slic} as superpixel generation technique. They first regroup superpixels by a spectral clustering algorithm into clusters and then perform a merging step using cluster similarities.
Another work from Yang et al. \cite{yang2016novelClustering01} presented a kernel fuzzy similarity measure which is used to cluster superpixels. A 10-dimensional texture-based feature vector is extracted to characterize each superpixel. After clustering, a k-means final step is applied to group clusters into final regions.
Yu and Wang\cite{yu2013context} proposed a graph-based coarse and fine merging strategy based on features extracted on superpixels including three Gestalt laws-inspired rules to model the superpixel context. 
In the same vein, the work of Oneata et al.\cite{oneata2014spatio} proposed a segmentation algorithm for action and event detection in videos. Similarity between regions is computed based on appearance, motion and geodesic spatio-temporal features. A hierarchical clustering with average linkage is then applied on the graph to produce the final segmentation result.

In all these works, we can note that superpixels to be merged are assumed to satisfy two main criteria: spatial adjacency and perceptual similarity. This implies that an efficient approach should be able to pick the most similar spatial neighbor of a superpixel in both local and global manners. In this paper, we present a region-growing segmentation approach that uses superpixels over-segmentation as input data. We propose a robust similarity measure between superpixels that efficiently combine content and border scale to provide an accurate measure. Moreover, we define a merging strategy to guide the region growing process through an adaptive merging criterion and a priority order between region aggregations.
\section{Proposed superpixel-based segmentation approach}\label{sect:method}
This work proposes an image segmentation approach based on iterative growing of superpixels. In particular, the proposed approach addresses the two key issues previously mentioned namely similarity measure and order of merging regions with neighboring (super)pixel issues, in region-growing techniques.

\subsection{Approach overview}\label{sect:overview}
\begin{figure}[h]
\begin{center}
\adjincludegraphics[width=0.8\linewidth]{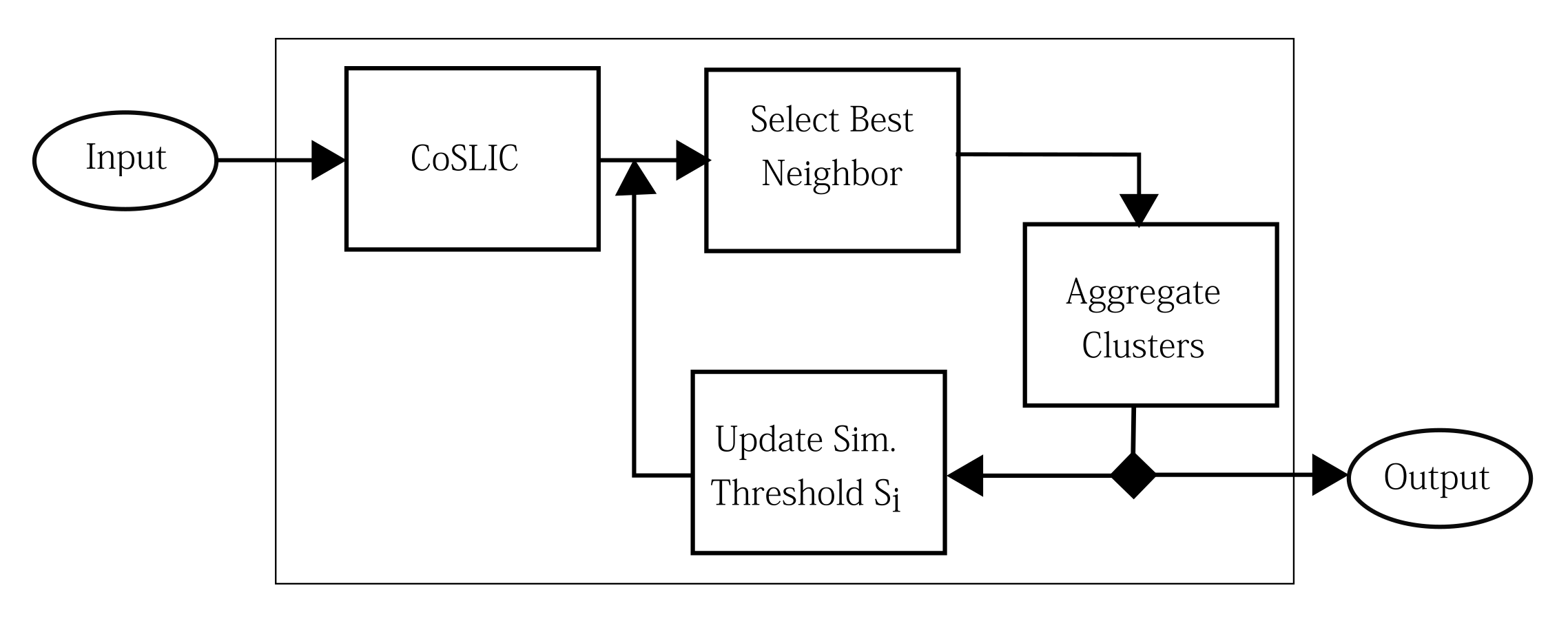} 
\caption {Global bloc overview of the proposed segmentation approach.} \label{fig:AppOverview} 
\end{center}
\end{figure}
As initialization, the proposed approach segments the source image into connected superpixels using the SLIC\cite{achanta2010slic} algorithm. SLIC is popular since it allows to generate superpixels with simple implementation, fast execution and quite good accuracy. However, it can be observed that it sometimes fails to perfectly adhere to image boundaries. Consequently, global contour constraints are applied (Sect.\ref{sect:ConstrainedSLIC}) on the SLIC algorithm to overcome this issue. 
Once an accurate superpixel decomposition is obtained, the iterative superpixels growing occurs. In particular, superpixels are grouped into regions according to two main criteria: mutual selection and global contour cross. Each region chooses the best merging candidate superpixel from its adjacent neighbors using a robust similarity measure (Sect.\ref{sect:ClustersSimilarity}) based on appropriate superpixel features (Sect.\ref{sect:SpClustersfeatures}). Afterwards, each couple of a merging region and its neighbor superpixel that have mutually chosen each other with enough similarity and which are not separated by a global contour are grouped together. This iterative merging process is repeated until we reach the final similarity threshold.
Note that at the first loop, each single superpixel is considered as a merging candidate region.
\\
Mutual selection and contour cross criteria ensure that each region is merged with the best superpixel within its neighborhood at a given iteration. It produces meaningful region aggregations in the first iterations but aggregations become random as regions continue to grow. 
Thus, we propose an adaptive similarity threshold to introduce a notion of priority between different possible region aggregations. In section~\ref{sect:ConsistencyBC}, we discuss this process in more detail. Indeed, the similarity threshold establishes the minimal similarity required for a region and a superpixel to be merged. It is initially set to the maximal value and is decreased or increased depending on the results of the previous iterations. When no aggregations occurred during the previous iteration, the similarity threshold is decreased because it is too high to allow aggregations. 
On the other hand, if aggregations occurred at previous iteration, the merged regions are more likely to have strong similarity than the unchanged ones. Thus, we increase the similarity threshold to focus the aggregations on these regions. This acts as seed regions, which are expanded to achieve segmentation.

This work presents three main contributions: (1) an extension of the SLIC algorithm for better boundaries adherence. (2) a robust two-term similarity measure based on content and border information for region aggregations. (3) a merging strategy which includes a priority order between region aggregations and an adaptive merge threshold criterion.
Algorithm \ref{algo} outlines the proposed approach while a bloc-based overview is shown in the figure~\ref{fig:AppOverview}.
\begin{algorithm}[t]
\linespread{1.2}\selectfont
\footnotesize
\KwData{$\mathcal{I}$,$\mathcal{S}_0$	\Comment*[r]{$\mathcal{I}$:Input image,$\mathcal{S}_0$:Stopping similarity}}
\KwResult{$\mathcal{S_P}$	\Comment*[r]{$\mathcal{S_P}$:superpixel set}}
\Begin{
$Gc := \textrm{Canny}(\mathcal{I})$\;
$\mathcal{S_P} := \textrm{CoSLIC}(\mathcal{I})$\;
$\mathcal{S}_{it} := 1 $	\Comment*[r]{$\mathcal{S}_{it}$:adaptive similarity}
$Ms := true $\;	

\While {$Ms = true \quad and \quad \mathcal{S}_{it} \geq \mathcal{S}_0$ }{
$Ms := false $\;
$ Mc := \emptyset$\;
\For{$P \in \mathcal{S_P}$}{
	$ N:=argmax_i\left(\textrm{Sim}(P,N_i)\right)$	\Comment*[r]{$\mathcal{N}(P)$:neighbors of $P$}
	$ \quad\quad \forall \quad N_i \in \mathcal{N}(P) \quad and \quad \textrm{Sim}(P,N_i)\geq \mathcal{S}_{it}$\;
	$ Mc := Mc \cup (P,N)$\;
}
\ForAll{$(P,Q) \in Mc$} {
\If{
$ (Mc(P) = Q \quad and \quad Mc(Q) = P)$
}
{
$Ms := true $\;
\If{$((P \cup Q) \cap Gc=\emptyset)$}{
$\mathcal{S_P}:=\textrm{Merge}(\mathcal{S_P},P,Q)$\;
$\mathcal{S}_{it}:=\alpha_{it}*\mathcal{S}_{it}$		\Comment*[r]{$\alpha_{it}$:sim update coefficient(\S\ref{sect:AdpativeCriterion})}
}
\Else{
$\mathcal{N}(P):=\mathcal{N}(P) - Q$\;
$\mathcal{N}(Q):=\mathcal{N}(Q) - P$\;
$\mathcal{S}_{it}:=\dfrac{1}{\alpha_{it}}*\mathcal{S}_{it}$\;
}
}
}
}
}
\caption{Superpixel-based Image Segmentation}\label{algo}
\end{algorithm}

\subsection{Contour-constrained SLIC (CoSLIC)}\label{sect:ConstrainedSLIC}
\begin{figure}[h]
\begin{center}
\adjincludegraphics[width=0.9\linewidth]{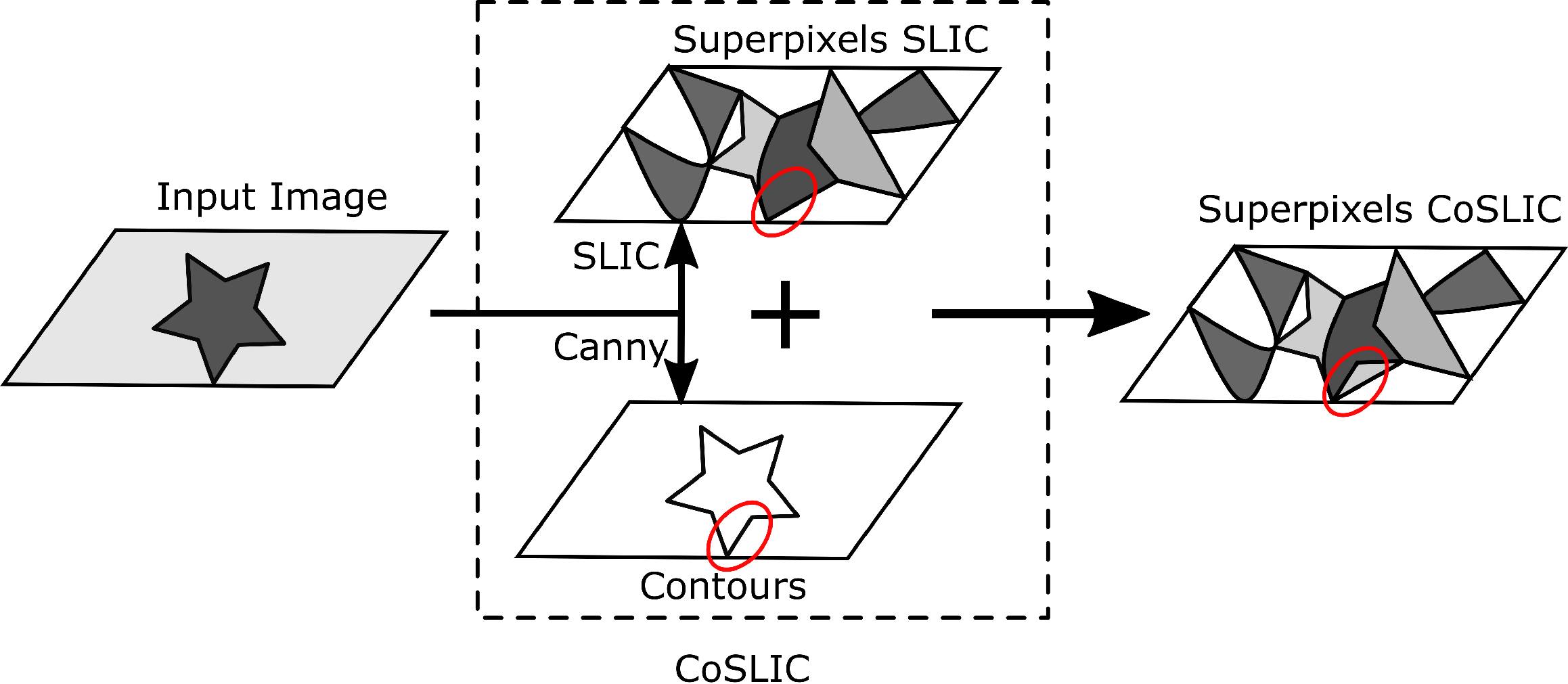} 
\caption {Global scheme of CoSLIC: an extension of SLIC\cite{achanta2010slic} using global contours for better boundary adherence} \label{fig:CoSLIC} 
\end{center}
\end{figure}

Despite its efficiency, SLIC algorithm often produces superpixels that overlap several regions on the input image boundaries. We propose to extend the SLIC algorithm by taking advantage of classical edge detectors to overcome this drawback. To this end, from the source image, global contour map are extracted using the Canny edge detector. Then, for each superpixel, a cross check with the global contours is performed. Superpixels that are crossed by global contours are split along the contours into smaller superpixels. \\
Given an input image $\mathcal{I}$, we denote by $\mathcal{S}_p$ the superpixel set generated by SLIC over $\mathcal{I}$, and $\mathcal{G}_C$ the contour map obtained by applying Canny edge detector on $\mathcal{I}$.
Our CoSLIC over-segmentation algorithm partitions each superpixel $P_i \in \mathcal{S}_p$ into $P_{i_j}$ sub-superpixels based on $\mathcal{G}_C$ such that the three following conditions are met
\begin{equation}
\begin{cases}
& (a) \quad P_{i_j} \cap \mathcal{G}_C = \emptyset, \forall i,j \\
& (b) \quad \bigcup_{j} P_{i_j} = P_i \\
& (c) \quad P_{i_j} \cap P_{i_k} = \emptyset, \forall j\neq k
\end{cases}
\end{equation} 
The first condition (a) ensures that no superpixel should be crossed by $\mathcal{G}_C$. The two last conditions define the obtained sub-superpixels $P_{i_j}$ as a partition of the parent superpixel $P_i$.
Figure\ref{fig:CoSLIC} shows the global scheme of the proposed SLIC extension :CoSLIC.

\subsection{Features extraction}\label{sect:SpClustersfeatures}
The process of superpixels growing towards accurate segmentation can be represented by a tree structure where nodes correspond to regions formed by groups of superpixels. This hierarchical tree is constructed in a bottom-up fashion. Two similar neighbor regions are grouped into one parent region node. Given an input image $\mathcal{I}$ initially partitioned into a set of superpixels $\mathcal{S_P}$. This definition of the region-growing approach can be expressed as a succession of ordered partitions $\Gamma_k$ of $\mathcal{I}$ at different levels $k$, going from $0$ to $K$. The root of the tree is denoted by the highest level $K$ and the leaves corresponding to initial superpixels set, denoted by level $0$. Thus, we have 
\begin{equation}
\forall \quad i,j \in [0,K], \quad
\begin{cases}
&i\neq j \implies \Gamma_i\neq \Gamma_j \\
&i<j \implies \lVert\mathcal{R}(\Gamma_i)\rVert \geq \lVert\mathcal{R}(\Gamma_j)\rVert
\end{cases}
\end{equation}
where $\mathcal{R(.)}$ defines the set of regions formed by superpixels aggregations and $\lVert.\rVert$ denotes the number of elements in a set.

\begin{figure*}
\centering \mbox{\includegraphics[width=0.8\textwidth]{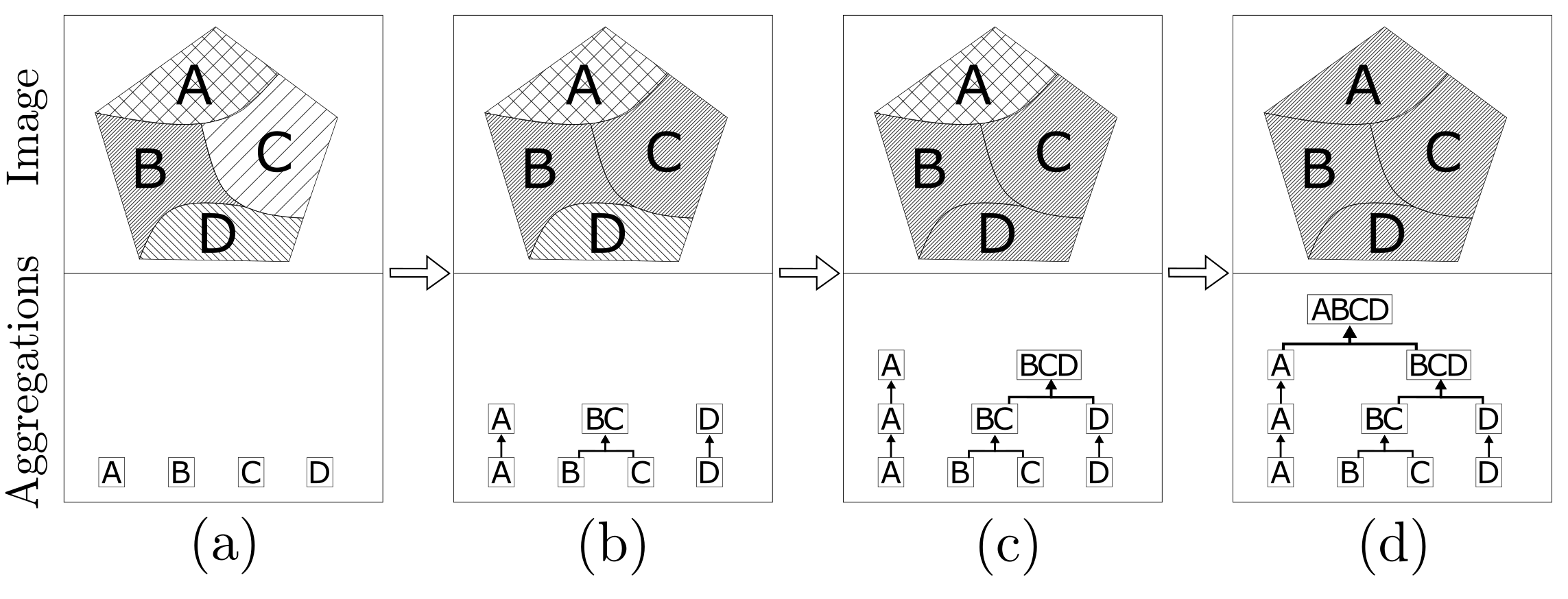}}
\caption {Hierarchical superpixel growing process using an image over-segmented in superpixels A,B,C and D. The iterative aggregations of superpixels is modeled through a hierarchical structure; Features are concatenated along such structure to reach a robust multi-scale region description.} \label{fig:HierarchicalCP}
\end{figure*}

Figure \ref{fig:HierarchicalCP} gives a visual illustration of this approach.
Superpixels are the only level to be directly related to the image in this hierarchical process. For this reason, superpixels are described by features extracted directly from image data. A superpixel $P$ is described by a vector of ten features $F_{P}=(f_{p}^0, \ldots,f_{p}^9)$ divided into intensity, texture and gradient features. Image is represented in the CIE Lab color space. The literature~\cite{tkalcic2003colour} established this color space to be correlated to the human visual perception. Moreover, classical vector distances like the Euclidean are proved to be visually meaningfull in the CIE Lab color space. Intensity feautres include mean, variance, skewness and a 10-bin histogram of the intensity values over each of the component L, a and b. Contrast, correlation, energy and entropy are extracted as texture features whereas histogram of orientations and histogram of magnitude are computed from the gradient of the image as summarized in Table~\ref{tab:Tblfeatures}.
\begin{table}[H]
\footnotesize
\caption{Features assigned to each single superpixel. Features are divided into intensity, texture and gradient features.} 
\label{tab:Tblfeatures}
\begin{center}       
\begin{tabular}{|l||l|l|}
\hline 
\rule[-1ex]{0pt}{3.5ex} 
{\bf Category} & {\bf Features } & {}\\
\hline \hline
\rule[-1ex]{0pt}{3.5ex}  
\multirow{4}{*}{Intensity} &  Mean & $f^0$ \\
							& Variance & $f^1$ \\
							& Skewness & $f^2$ \\
							& 10-bin Histogram & $f^3$\\
\hline
\rule[-1ex]{0pt}{3.5ex}  
\multirow{4}{*}{Texture} 	& Contrast & $f^4$ \\
							& Correlation & $f^5$\\
							& Energy & $f^6$\\
							& Entropy & $f^7$ \\
\hline
\rule[-1ex]{0pt}{3.5ex} 
\multirow{3}{*}{Gradient} 	& 10-bin Histogram of Orientations & $f^8$ \\
    						& 10-bin Histogram of Magnitude & $f^9$ \\
\hline
\end{tabular}
\end{center}
\end{table} 
Regions are characterized by a concatenation of their direct descendant regions in the hierarchical model. This describes the sequence of regions that are used to form the region and gives a multi-scale representation of the region under consideration. This idea \cite{conze2016} has been initially used to represent a leaf node by a sequence of its ascendants. In the proposed approach, this idea is applied in a bottom-up way to suit our approach and keep the same outcome. Thereby, each region has a multi-level representation that contains the description of all its ordered descendants according to their level in its hierarchy. In this description space, the region $BC$ in fig.~\ref{fig:HierarchicalCP}-b will be characterized by $F_{BC}=(f_B,f_C)$ where $F_B=(f_{B}^0, \ldots,f_{B}^9)$ and $F_C=(f_{C}^0, \ldots,f_{C}^9)$ describing respectively the superpixels $B$ and $C$.
\subsection{Regions similarity measure}\label{sect:ClustersSimilarity}
This work presents a region-growing approach. A region $R\in\Psi_r$ is defined, from a set of superpixels $\mathcal{S}_p$, as a non-empty superpixel cluster. Equation~\ref{Eqn:RegionSet} defines the set of regions, $\Psi_r$.
\begin{equation}\label{Eqn:RegionSet}
\begin{array}{l}
 	\Psi_r=\{x:x\subseteq\{\mathcal{S}_p - \emptyset\}\}
\end{array}
\end{equation}
\\Similarity measure between regions is a key component of a region-growing segmentation. The proposed similarity measure computes the difference both at region and border levels, giving at the same time a global and a local view to the comparison. The idea is that two regions to be merged must have similar content and a smooth common border meaning that there is no abrupt change from one region to the other. 
Therefore, the similarity between regions is defined from their content but also from their common border. The content similarity is computed as a comparison of the two regions, whereas the border similarity is defined based on the similarity between the connected superpixels that form the border from each region. Thus the general form of the similarity measure between two regions, $R_i,R_j\in\Psi_r$, can be expressed by Equation \ref{eqn:SimilariltyMeasure}.
\begin{equation}\label{eqn:SimilariltyMeasure}
Sim(R_i,R_j)=f(Sim_C(R_i,R_j),Sim_B(R_i,R_j))
\end{equation} 
where $Sim_C(R_i,R_j)$ and $Sim_B(R_i,R_j)$ denote respectively the content and the border similarity between $R_i$ and $R_j$.

\subsubsection{Content-based similarity}\label{sect:ContentSimilarity}
First, the content similarity between $R_i$ and $R_j$, represented by their respective multi-scale features vectors $f_i$ and $f_j$ is defined.
Considering only the ${\ell}^\textrm{\scriptsize th}$ feature, an intermediate content similarity between $R_i$ and $R_j$ is also defined as:
\begin{equation}
Sim_C^{\ell}(R_i,R_j)=\mu({\chi}^2(f_i^{\ell},f_j^{\ell})),
\end{equation}
where :
\begin{itemize}
\item[-] $\mu(x)=\exp\left(\dfrac{1}{2}*\left(\dfrac{x}{\sigma}\right)^2\right)$ is the zero-centered Gaussian fuzzy membership function used here to normalize the similarity value and $\sigma$ its standard deviation value.
\item[-] The ${\chi}^2$ function comes from the ${\chi}^2$ test-statistic~\cite{snedecor1967statistical} where it is used to test the fit between a distribution and observed frequencies. It is defined, for two vectors $x=(x_1,\ldots, x_n) \textrm{ and } y=(y_1,\ldots, y_n)$, as ${\chi}^2(x,y)=\sum_{i=1}^n\left(\dfrac{(x_i-y_i)^2}{(x_i+y_i)}\right)$.
\end{itemize} 
Along all features, the vector of the intermediate feature-level similarities can be expressed as:
\begin{equation}
\begin{aligned} 
	S^{\ell} &=
         \begin{bmatrix}
           Sim^{\ell}_C(f^0_i,f^0_j)) \\
           \vdots \\
           Sim^{\ell}_C(f^9_i,f^9_j))
         \end{bmatrix}
         =
    	\begin{bmatrix}
           \mu({\chi}^2(f^0_i,f^0_j)) \\
           \vdots \\
           \mu({\chi}^2(f^9_i,f^9_j))
         \end{bmatrix}
         \\&=
         \begin{bmatrix}
           S_{0} \\
           \vdots \\
           S_{9}
         \end{bmatrix} 
         , \texttt{$S_i \in [0,1], i=(0,\ldots,9)$}
  \end{aligned}
\end{equation}
The final content similarity between $R_i$ and $R_j$, $Sim_{C}(R_i,R_j)$, is defined as an average of some statistical values of the distribution of the intermediate similarities between the two regions by going through all the features. It is defined by:
\begin{equation}
Sim_{C}(R_i,R_j)=\dfrac{1}{4}\left(S_{max}+S_{min}+S_{mean}+S_{var}\right)
\end{equation}
where:
\begin{itemize}
\item[-]  $S_{max}=max_{\ell}(S^{\ell})$, $S_{min}=min_{\ell}(S^{\ell})$ are the extrema of the feature-level similarities between the two regions and
\item[-] $S_{mean}=mean_{\ell}(S^{\ell})$, $S_{var}=var_{\ell}(S^{\ell})$ provide the measure of central tendency and dispersion of those similarities as the mean and the variance, respectively.
\end{itemize} 
A region is described by a range of features with different dynamics. 
Thus, by normalizing those intermediate feature-level similarities, we ensure that the calculated similarity will not be biased by the dynamics of the features. Furthermore, we experimentally observe that such statistics give better similarity than a classical distance, like the euclidean, between the intermediate similarities.
\subsubsection{Border-based similarity}\label{sect:BorderSimilarity}
\begin{figure}
\begin{center}
\includegraphics[scale=0.5]{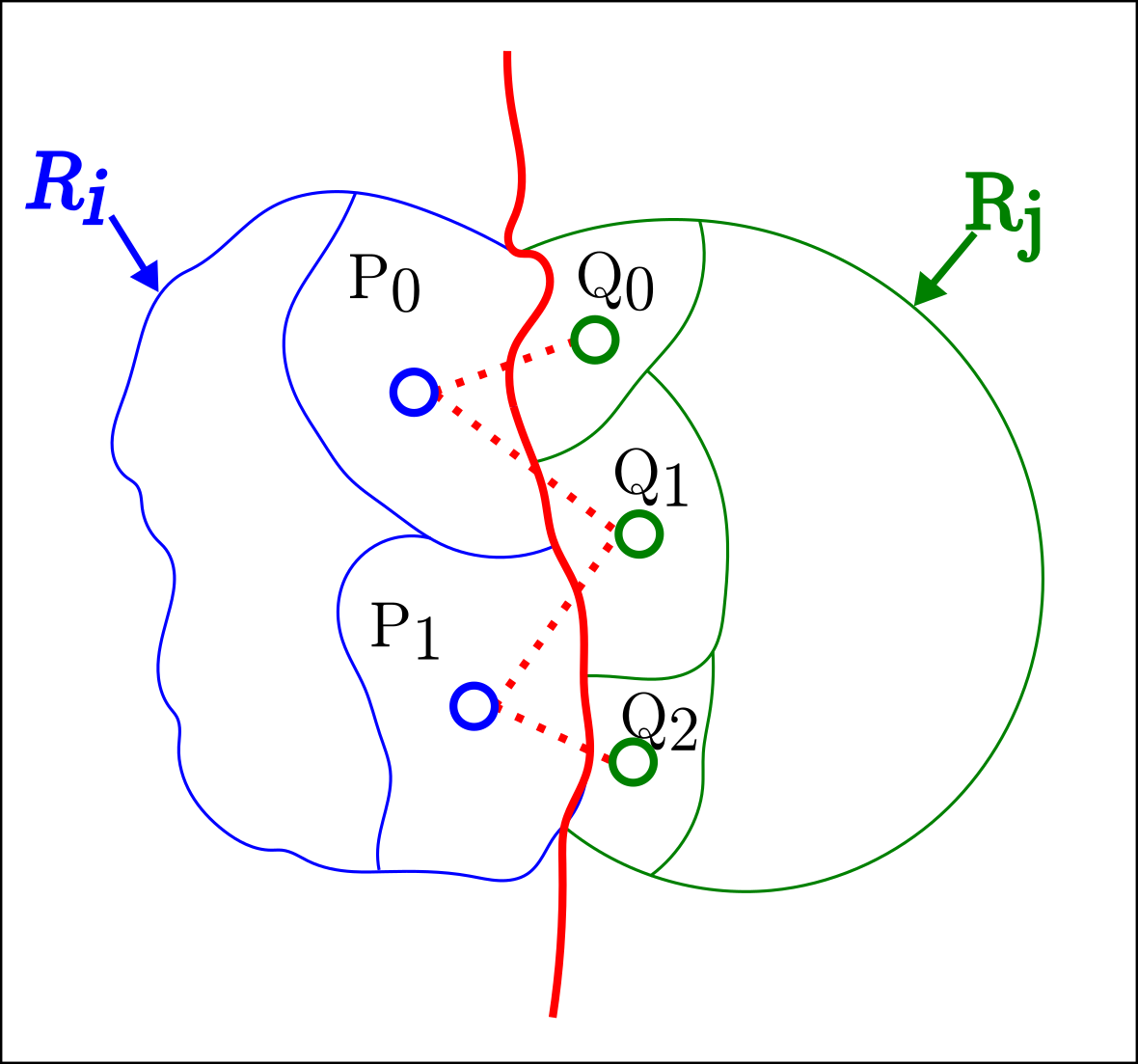}
\caption{Border similarity measure: the two regions $R_i$ and $R_j$ are separated by 4 couples of neighboring superpixels. The border similarity is computed as $Sim_B(R_i, R_j)=\dfrac{1}{4}[Sim_C(P_0,Q_0)+Sim_C(P_0,Q_1)+Sim_C(P_1,Q_1)+Sim_C(P_1,Q_2)]$}
\label{fig:SimBorder}
\end{center}
\end{figure}

The border similarity is designed to prevent similar adjacent regions from different objects to be merged. Two neighboring regions $R_i$ and $R_j$ are assumed separated by a border of $N$ couple of superpixels that we will denote by $P_i$ for superpixels that belong to $R_i$ and $Q_j$ for those from $R_j$. Thus, Equation~\ref{eqn:BorderSimilarity} gives the expression of the border similarity measure between $R_i$ and $R_j$. An illustrative example is shown in the figure~\ref{fig:SimBorder}.
\begin{equation}\label{eqn:BorderSimilarity}
Sim_B(R_i,R_j)=\dfrac{1}{N}\left(\sum_{P_i \in \mathcal{N}(Q_j)}^NSim_{C}(P_i,Q_j)\right)
\end{equation}
In other words, $Sim_B(R_i,R_j)$ is considered as the average of the similarities between each couple of superpixels along the border separating $R_i$ and $R_j$.

\subsubsection{Similarity measure}\label{sect:GlobalSimilarity}
Finally, the similarity between the two regions, $Sim(R_i,R_j)$, is defined as a weighted combination of the content and the border similarities as expressed by Equation~\ref{eqn:SimilarityTotal}. The content weighting coefficient, defined by $\omega_C=\sqrt{\dfrac{min\left(\lVert R_i \rVert,\lVert R_j \rVert\right)}{max\left(\lVert R_i \rVert,\lVert R_j \rVert\right)}} $ is added in order to balance the regions similarity accordingly with respect to their respective size. Also, given $\mathcal{C}_i$ (resp. $\mathcal{C}_j$) as the circumference of $R_i$ (resp. $R_j$) and $\beta$ the length of their common border in pixels, we defined the weight of $Sim_B(R_i,R_j)$ as $\omega_B=\sqrt{\dfrac{\beta}{2*\mathcal{C}_i*\mathcal{C}_j}*(\mathcal{C}_i+\mathcal{C}_j)}$ which corresponds to the percentage of the common border between the two regions. 
\begin{equation}\label{eqn:SimilarityTotal}
Sim(R_i,R_j)=\omega_C*Sim_C(R_i,R_j)+\omega_B*Sim_B(R_i,R_j)
\end{equation}

\noindent The content coefficient favors grouping of regions of similar sizes, while the border coefficient favors the grouping of regions that share the largest boundary. This acts as a regulator and prevents a group of regions from growing faster than others but on a regular basis. 

\subsection{Regions merging strategy}\label{sect:ConsistencyBC}
\subsubsection{Merging procedure}\label{sect:MergingProcedure}
The proposed merging procedure is inspired by the agglomerative clustering algorithm~\cite{ward1963hierarchical} for its conceptual simplicity and flexibility. Region-growing algorithms are inherently dependent on the order of processing of the image pixels\cite{mehnert1997improved}. Merging order strongly affects the convergence of the approach because if two regions are wrongly merged then this error will be spread to the next steps. To tackle this issue, we propose a merging strategy that minimizes errors at each step. In fact, at a given iteration, every region makes a unique choice of merging candidates from its spatially neighboring regions. This selection ensures local optimal choice. Once each region has made its choice, a validation step performs a global optimization check to remove all regions that are not mutually chosen by each other or which are separated by a global contour. Thus, we avoid any conflicts and guarantee maximum coherence to the grouping step. 
Algorithm~\ref{alg:MergingProcedure} summarizes the main steps of the proposed merging procedure. 

\begin{algorithm}[tp]
\linespread{1.2}\selectfont
\footnotesize
\tcc{Best local configuration}
\For{$P \in \mathcal{S_P}$}{
	$ N:=argmax_i\left(\textrm{Sim}(P,N_i)\right)$
	$ \quad\quad \forall \quad N_i \in \mathcal{N}(P) \quad and \quad \textrm{Sim}(P,N_i)\geq \mathcal{S}_{it}$\;
	$ \mathcal{N}^*(P) := N$ 	\Comment*[r]{$\mathcal{N}^*(P)$: best neighbor of $P$}
}

\tcc{Best global configuration}
\ForAll{$P \in \mathcal{N}^* \quad|\quad \mathcal{N}^*(P)=Q$} {
\If{
$ (\mathcal{N}^*(P) = Q \quad and \quad \mathcal{N}^*(Q) = P)$
}
{
\If{$((P \cup Q) \cap GlobalContours=\emptyset)$}{
$\mathcal{A}^*(P)=Q$ 	\Comment*[r]{$\mathcal{A}^*(P)$: aggregation neighbor of $P$}
$\mathcal{A}^*(Q)=P$\;
$\mathcal{S}_{it}:=\alpha_{it}*\mathcal{S}_{it}$\;
}
\Else{
$\mathcal{N}(P):=\mathcal{N}(P) - Q$\;
$\mathcal{N}(Q):=\mathcal{N}(Q) - P$\;
$\mathcal{S}_{it}:=\dfrac{1}{\alpha_{it}}*\mathcal{S}_{it}$\;
}
}
}

\tcc{Aggregation priority order}
\ForAll{$P \in \mathcal{A}^* \quad|\quad \mathcal{A}^*(P)=Q$}{
\If{$Sim(P,Q)>\mathcal{S}_{it}$}{
$\mathcal{S_P}:=\textrm{Merge}(\mathcal{S_P},P,Q)$\;
}
}
\caption{Region merging}\label{alg:MergingProcedure}
\end{algorithm}

To sum up, the merging procedure consists of three consecutive steps.
Given a region $R_i$, we first choose the best merging configuration by identifying its best neighboring region $R_j$, using the following measure:
$\mathrm{Sim}(R_i,R_j) = max\left(\mathrm{Sim}(R_i,N_i)\right)$ 
$ \forall \quad N_{i} \in \mathcal{N}(R_i)$.
Second, the image global best merge configuration is made from the best configuration of all merging regions pairs. To this end, we suppose that 
$\mathcal{N}^{*}(R_i) = R_j \quad \text{and} \quad \mathcal{N}^{*}(R) = R_j \quad \text{, then we have} \quad 
\mathrm{Sim}(R_i,R_j)<\mathrm{Sim}(R,R_j) \implies \mathcal{N}^{*}(R_i) = \emptyset$  where $\mathcal{N}^{*}(.)$ indicates the best region neighbor. 
Third, we ensure priority between region aggregations by validating preferential aggregations only. An aggregation is preferential when the regions have a similarity greater than $\mathcal{S}_{it}$.

\subsubsection{Adaptive merging criterion}\label{sect:AdpativeCriterion}
In classical region-growing image segmentation techniques, the compared regions belong to the same merging iteration. There is no simultaneous considerations of regions generated from different iterations. However, since the size and content of regions change through iterations, a given similarity value may not express the same visual similarity at different iterations. As an example, considering color similarity as merging criterion, a given similarity value will not correspond to the same visual similarity when comparing homogeneous small regions (early iterations) and heterogeneous big regions (after some iterations). This means that the merging criterion value must not be a fixed value, but must be updated to remain consistent through iterations. 
This idea is introduced in the proposed approach by using an adaptive similarity threshold which is updated accordingly to the results of previous iteration. Basically, this threshold is decreased when no aggregations occurred at the previous iteration otherwise it is increased. In the case where no aggregation occurs at the previous iteration, the similarity threshold is too high so it is decreased to allow aggregations at the next iteration. In the opposite case, the similarity of the newly formed regions with their neighborhood will increase as their content gathers together the contents of the two regions that formed them. Thus, the current threshold is increased in order to filter irrelevant neighbor regions. The similarity threshold is updated according to the iteration coefficient $\alpha_{it}$. By applying this adaptive similarity threshold, we ensure best region aggregations to happen first, which allows us to impose a priority order to those aggregations.
\begin{equation}
\alpha_{it}=1+\dfrac{\lVert mergedRegions_{i-1}\rVert}{\lVert candidateRegions_{i-1}\rVert}
\end{equation}
where $\lVert.\rVert$ denotes the size of the set under consideration and $i$ is the rank of the current iteration.

\section{Experiments and results}\label{sect:results}
To validate the proposed approach, expriments are conducted on the Berkeley Segmentation Dataset and Benchmark (BSDS500)\cite{martin2001database}. This database consists of natural images with five different human groundtruth (GT) segmentations for each.
The proposed approach is compared to other image segmentation approaches on 100 images, randomly selected from the BSDS500. As the global similarity between two regions is normalized, we empirically set the value of the stopping similarity $\mathcal{S}_0$ to $0.4$ during all the experiments. The main results are presented and discussed below.
\subsection{Evaluation}\label{resEvaluation}
We consider the six following segmentation quality assessment criteria to compare the proposed approach with state-of-the-art methods. The Boundary Recall and the Under-segmentation error are used to evaluate the CoSLIC over-segmentation results whereas the last four criteria are used to assess final segmentation results.

\begin{enumerate}
\item Boundary Recall (BR) measures the fraction of groundtruth edges that fall within at least one superpixel boundary, with a tolerance distance $\delta$ (usually set to $2$ $pixels$).
\item Under-segmentation Error~\cite{neubert2012UnderSegmentation} (UE) compares superpixels areas to measure in what extend superpixels flood over the groundtruth segment borders. 

\item Probabilistic Rand Index \cite{unnikrishnan2007toward} (PRI) which is a measure of similarity between two data clusters according to their labeling. In image segmentation, it measures the proportion of pixels that have the same labels compared to groundtruth segmentation. 
\item Variation of Information \cite{meilua2007comparing} (VoI) which gives the dissimilarity between two clusterings results. Based on a conditional entropy expression, VoI measures the amount of randomly clustered pixels in the segmentation with no clues in the groundtruth. 
\item Boundary Displacement Error \cite{freixenet2002yet} (BDE) which measures the average displacement error of one boundary pixel and the closest boundary pixels in the groundtruth segmentation. 
\item Global Consistency Error \cite{martin2001database} (GCE) measuring the extent to which one segmentation is a
refinement of the other.
\end{enumerate}
The higher BR value, the better the over-segmentation unlike UE whose high values denote low over-segmentation quality.
A good segmentation algorithm results should have high PRI values as well as low VoI, BDE and GCE.
\\
In order to evaluate the proposed approach, obtained results are compared with those of four state-of-the-art image segmentation approaches: Normalized Cut~\cite{shi2000normalized}, MeanShift~\cite{comaniciu2002mean}, CTM~\cite{yang2008unsupervised} and HFEM~\cite{yin2017unsupervised}. In Normalized Cut, an image is modeled by an un-directed graph. Segmentation is achieved by partitioning the graph while minimizing the overall cuts. This approach focuses on the similarity between image segments. The Mean Shift approach is based on the application of a non parametric density estimator to image segmentation. Adjacent pixels are grouped together according to a similarity criterion to form the final segmentation regions. 
The CTM approach transforms natural-image segmentation into multivariate mixed texture data clustering problem. Texture features are computed by using either 2D texture filter banks or simple fixed-size windows. The authors show that the mixture can effectively be segmented by a simple agglomerative clustering algorithm derived from a lossy data compression approach by minimizing the overall coding length of the feature vectors.
The Un-supervised Hierarchical Image Segmentation through Fuzzy Entropy Maximization (HFEM) is an un-supervised multi-level segmentation algorithm. The algorithm maximizes both fuzzy 2-partition entropy and segmentation smoothness through is a bi-level segmentation operator that uses binary graph cuts.
\\
As in the BSDS500 5 GT images are provided for every image, our segmentation results are evaluated by comparing each segmented image with all the GT images separately. The resulting evaluation value for an image is the average of the groundtruth-wise comparison.

\subsection{Discussion}\label{resDiscussion}
\subsubsection{Over-segmentation}\label{resDiscussionOverseg}
\begin{figure*}[!b]
\begin{center}
\adjincludegraphics[width=0.95\textwidth]{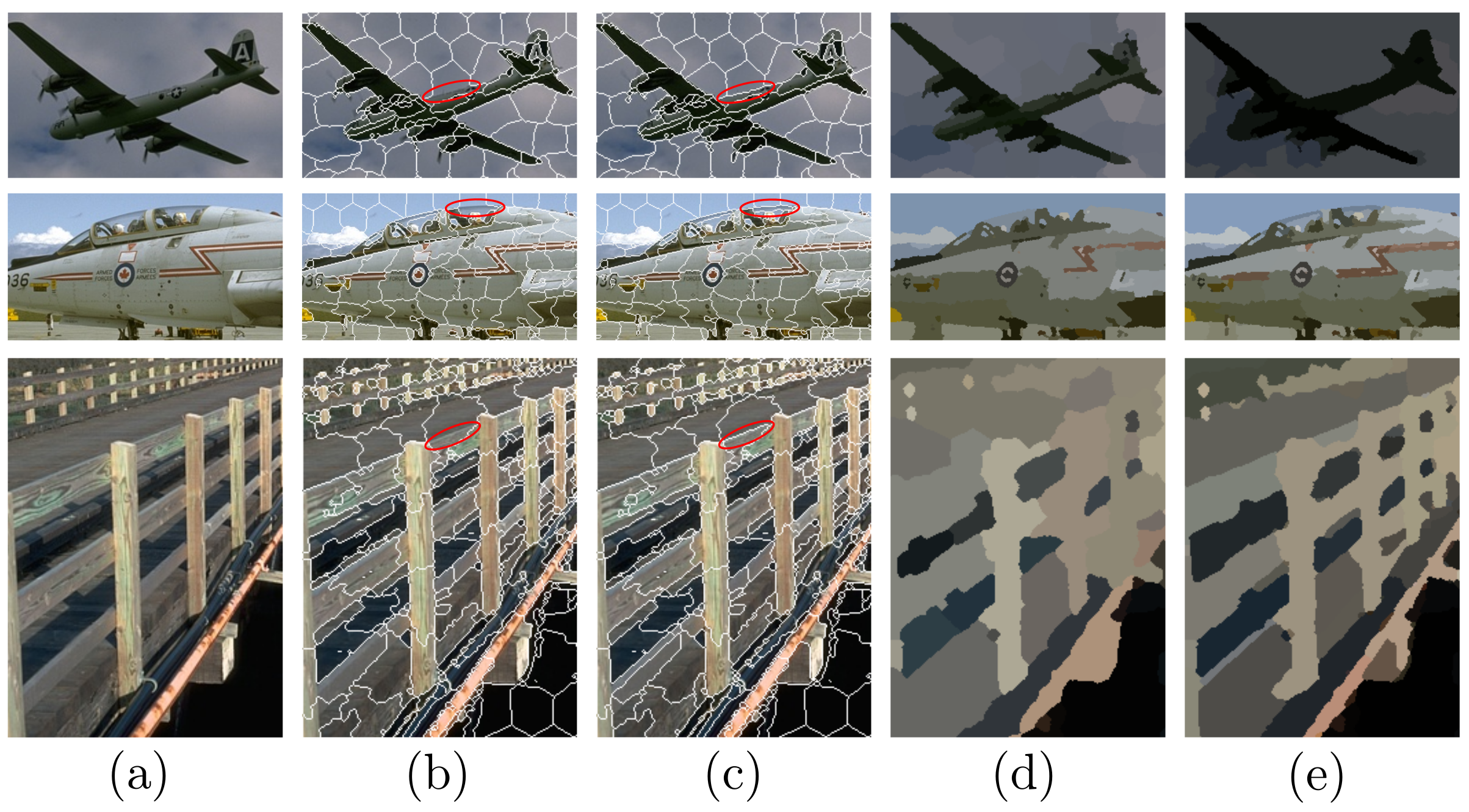}
\caption {Some results of the proposed extension of SLIC\cite{achanta2010slic} using global contours to improve boundary adherence. (a) original image. The SLIC superpixels are given in (b) and the corrected result using global contours is provided by (c). Global contours allow to recover some boundaries omitted by SLIC. (d) and (e) give segmentation results through superpixel-based region growing based on SLIC and CoSLIC superpixels respectively.} \label{fig:SLICGlobalContours} 
\end{center}
\end{figure*}

\begin{figure*}[!htb]
\begin{center}
\includegraphics[width=0.95\linewidth]{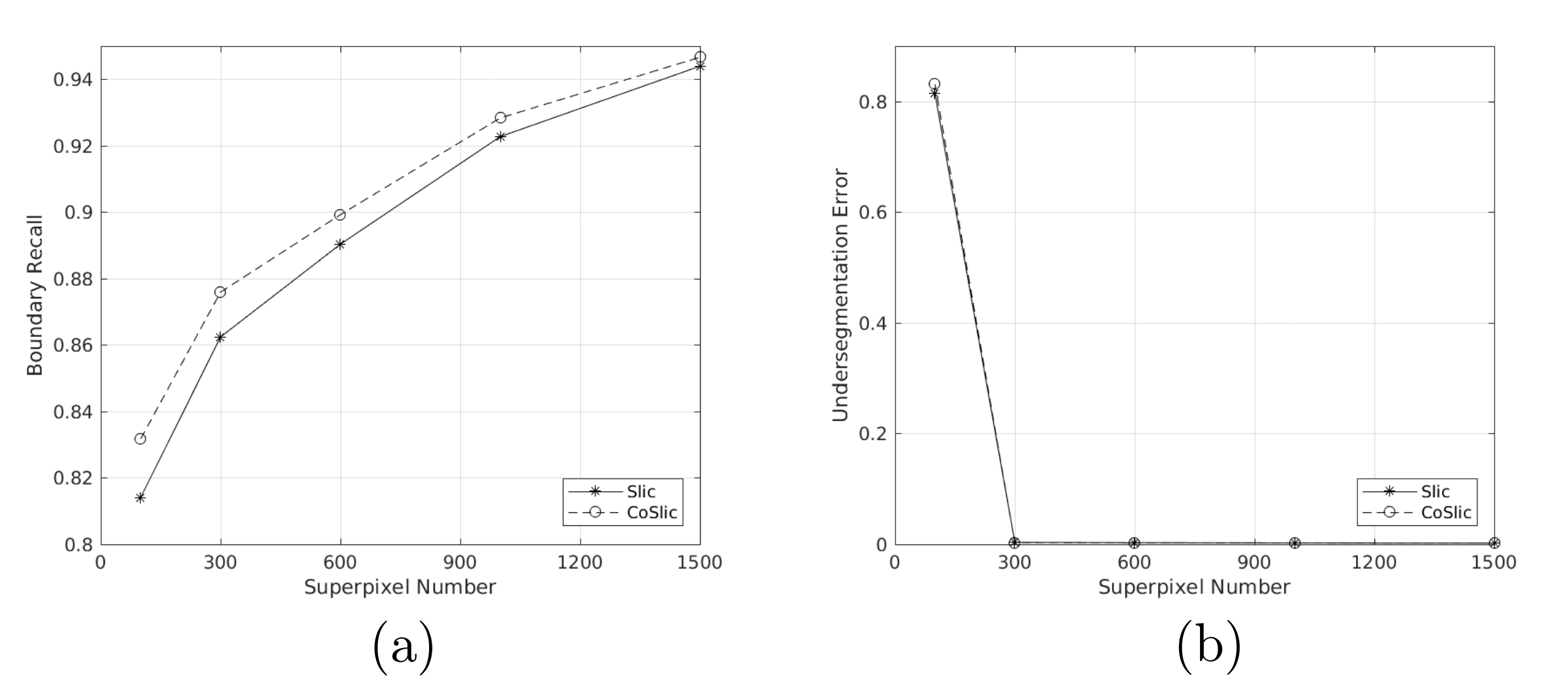}
\caption {Over-segmentation results evaluation: (a) Boundary Recall and (b) Under-segmentation Error curves for SLIC and CoSLIC.} \label{fig:CoSLICBR-UE} 
\end{center}
\end{figure*}

The purpose of this work is to segment images through superpixel-based region-growing.
Furthermore, with regard to the groundtruth, CoSLIC algorithm performs better than SLIC in many cases as shown in figure~\ref{fig:SLICGlobalContours}. CoSLIC outperforms SLIC in images containing neighboring objects separated by a thin borders because SLIC fails to detect the transitions between objects. As shown in Fig.~\ref{fig:CoSLICBR-UE}, the performance of proposed over-segmentation algorithm is higher under the boundary recall metric than SLIC. This outcome is excepted as CoSLIC is designed to provide better boundary adherence than SLIC. However, the results under the under-segmentation error are quite similar. 
The proposed approach is based on superpixels. Figure~\ref{fig:SLICGlobalContours} presents some results of the proposed extension of the SLIC algorithm. The interest of the added contour correction appears clearly. Segmentation results from an initial over-segmentation with SLIC (\ref{fig:SLICGlobalContours}-d) and from CoSLIC (\ref{fig:SLICGlobalContours}-e) show drastical changes on final results when few boundaries are not detected by the SLIC algorithm.

\subsubsection{Final Segmentation}\label{resDiscussionSeg}
\begin{figure*}
\begin{center}
\includegraphics[width=0.95\textwidth]{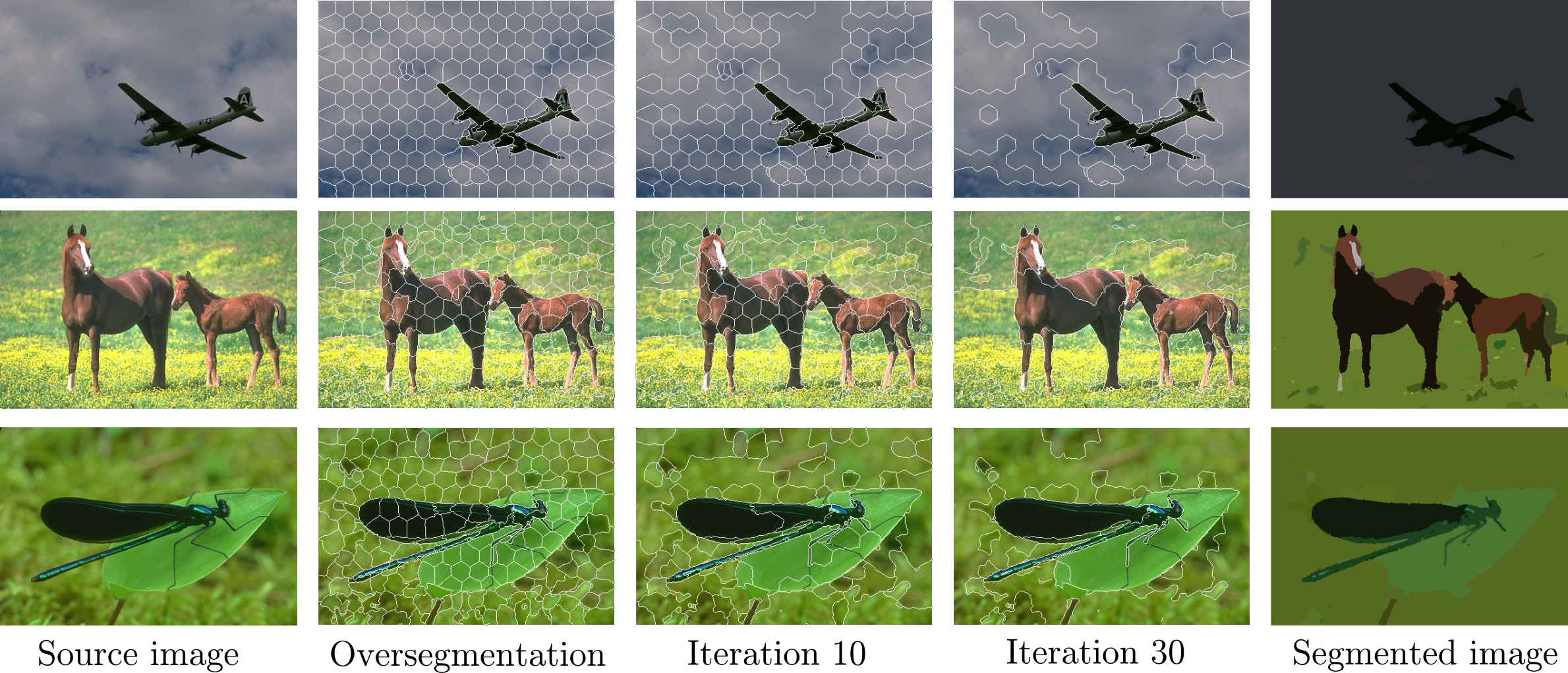}
\caption{Segmentation process: intermediate segmentation results.} \label{fig:results02} 
\end{center}
\end{figure*}

\begin{figure*}
\begin{center}
\includegraphics[width=0.95\textwidth]{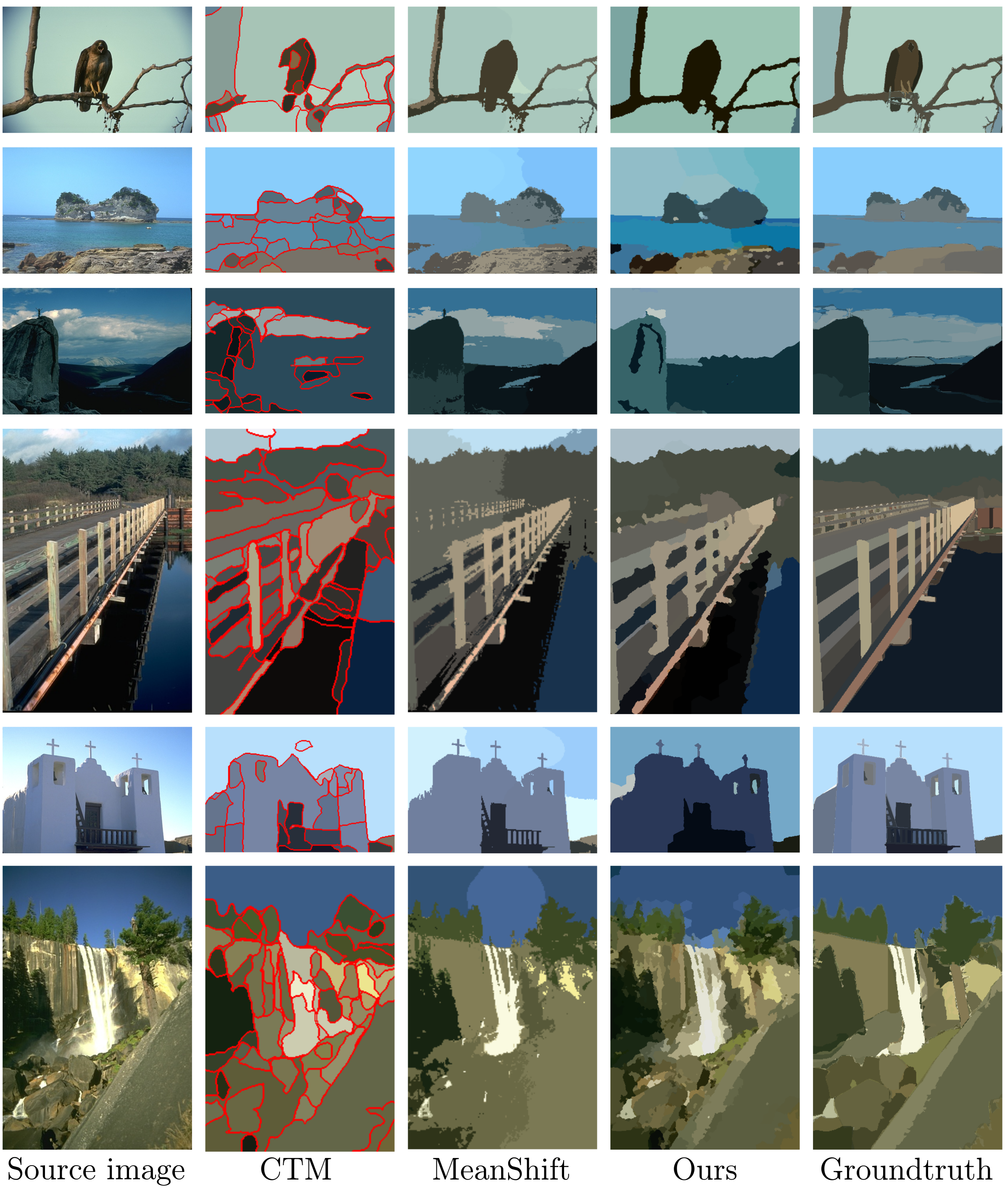}
\caption{Segmentation results: some results of our segmentation approach along with those of CTM and MeanShift on the BSDS500 Image Dataset} \label{fig:results01} 
\end{center}
\end{figure*}

\def\arraystretch{1.5}
\begin{table}[H]
\footnotesize
\caption{Summary of the performance evaluation results of our approach, NCut, CTM, HFEM and MeanShift on BSDS500. The best segmentation should have the highest PRI while its VoI, BDE and GCE should be the lowest.  Best results are in bold} 
\label{tab:TblEvaluation}
\begin{center}       
\begin{tabular}{l||c|c|c|c|}
	\cline{2-5} & {PRI $\uparrow$} & {VoI $\downarrow$} & {BDE $\downarrow$} & {GCE $\downarrow$} \\ \hline \hline
	\multicolumn{1}{|l||}{NCut~\cite{shi2000normalized}} & 0.73931 & 2.9139 & 17.1560 & 0.2232 \\ \hline 
	\multicolumn{1}{|l||}{CTM~\cite{yang2008unsupervised}} & \textbf{0.7796} &  6.2187 &  19.1981 & 0.3647 \\ \hline
	\multicolumn{1}{|l||}{HFEM~\cite{yin2017unsupervised}} & 0.7769 &  \textbf{2.3067} & 10.6700 & \textbf{0.2215} \\ \hline
	\multicolumn{1}{|l||}{MeanShift\cite{comaniciu2002mean}} & 0.7769 & 4.3173 & 13.1616 & 0.5811 \\ \hline 
	\multicolumn{1}{|l||}{Proposed approach} & 0.7627 & 3.8036 & \textbf{10.1594} & 0.4484 \\ \hline
\end{tabular}
\end{center}
\end{table} 
In the segmentation task, we particularly focus on the similarity measure and the merging strategy in the segmentation process.\\
The merging strategy is designed to guide the merging iterations during segmentation. This process effectively manage to first merge visually similar neighbors, as can be seen in Fig~\ref{fig:results02} through iterations.\\
Considering the results given in Table~\ref{tab:TblEvaluation}, the proposed approach presents overall competitive results with respect to compared approaches. In addition, we note that the proposed approach has the best result in terms of BDE. Additionally to visual results provided in Fig.~\ref{fig:results01}, it confirms the quality of the initial superpixels but also of the measure of similarity. However, in some cases as can be seen in images presented in Fig.\ref{fig:results01}-image 6, our approach tends to over-segment the images. This behavior may be caused by the final similarity value that stops aggregations of clusters. This value is determined empirically and therefore may not be suitable for all images.\\
The performances according to PRI criteria tend to conclude that our approach is superior to NCut and comes slightly under the other approaches. This criterion measures the quality of the pixel classification compared to the groundtruth. Indeed, the proposed approach does not explicitly include constraints on the labeling of pixels but still manages this through the aggregation order introduced in the merging strategy. \\
The HFEM has the best results in both VoI and GCE. The proposed approach outperforms the MeanShift in terms of GCE and performs better than MeanShift and CTM in terms of VoI. In fact, our approach, unlike the other approaches, does not consider a global view at the image level but only at superpixel cluster level.

\section{Conclusion}\label{sect:conclusion}
This paper describes an image segmentation approach based on iterative superpixel aggregation. First, an extension for the SLIC algorithm is proposed in order to provide better boundary adherence. Second, we proposed a robust similarity measure for comparing regions using both global multi-scale and common border criteria. This measure integrates border information to prevent merging overlapped regions. Third, a merging strategy is designed to control region aggregations through iterations using an adaptive similarity threshold. This strategy ensures that aggregations occur in the order of decreasing similarity. We confirmed the effectiveness of the proposed contributions through comparisons with well-established segmentation approaches using the BSDS500 image dataset. For further research, the proposed superpixel-based region-growing method can be easily extended to semantic segmentation since aggregated superpixels can provide a powerful high-level semantic description.
\bibliographystyle{abbrv}  
\bibliography{report}   



\end{spacing}
\end{document}